\documentclass[preprint,12pt,authoryear]{elsarticle}

\usepackage{amssymb}
\usepackage{amsmath}
\usepackage{booktabs} 

\journal{Nuclear Physics B}

\begin{document}

\begin{frontmatter}



\title{Systematic Weight Evaluation for Pruning Large Language Models: Enhancing Performance and Sustainability}


\author[inst1]{Ashhadul Islam}

\affiliation[inst1]{organization={College Of Science \& Engineering, Hamad Bin Khalifa University},
            addressline={Education City}, 
            city={Doha},
            postcode={34110}, 
            country={Qatar}}

\author[inst1]{Samir Brahim Belhaouari}
\author[inst1]{Amine Bermak}


\begin{abstract}
The exponential growth of large language models (LLMs) like ChatGPT has revolutionized artificial intelligence, offering unprecedented capabilities in natural language processing. However, the extensive computational resources required for training these models have significant environmental implications, including high carbon emissions, energy consumption, and water usage. This research presents a novel approach to LLM pruning, focusing on the systematic evaluation of individual weight importance throughout the training process. By monitoring parameter evolution over time, we propose a method that effectively reduces model size without compromising performance. Extensive experiments with both a scaled-down LLM and a large multimodal model reveal that moderate pruning enhances efficiency and reduces loss, while excessive pruning drastically deteriorates model performance. These findings highlight the critical need for optimized AI models to ensure sustainable development, balancing technological advancement with environmental responsibility.
\end{abstract}

\begin{graphicalabstract}
\includegraphics[width=\textwidth]{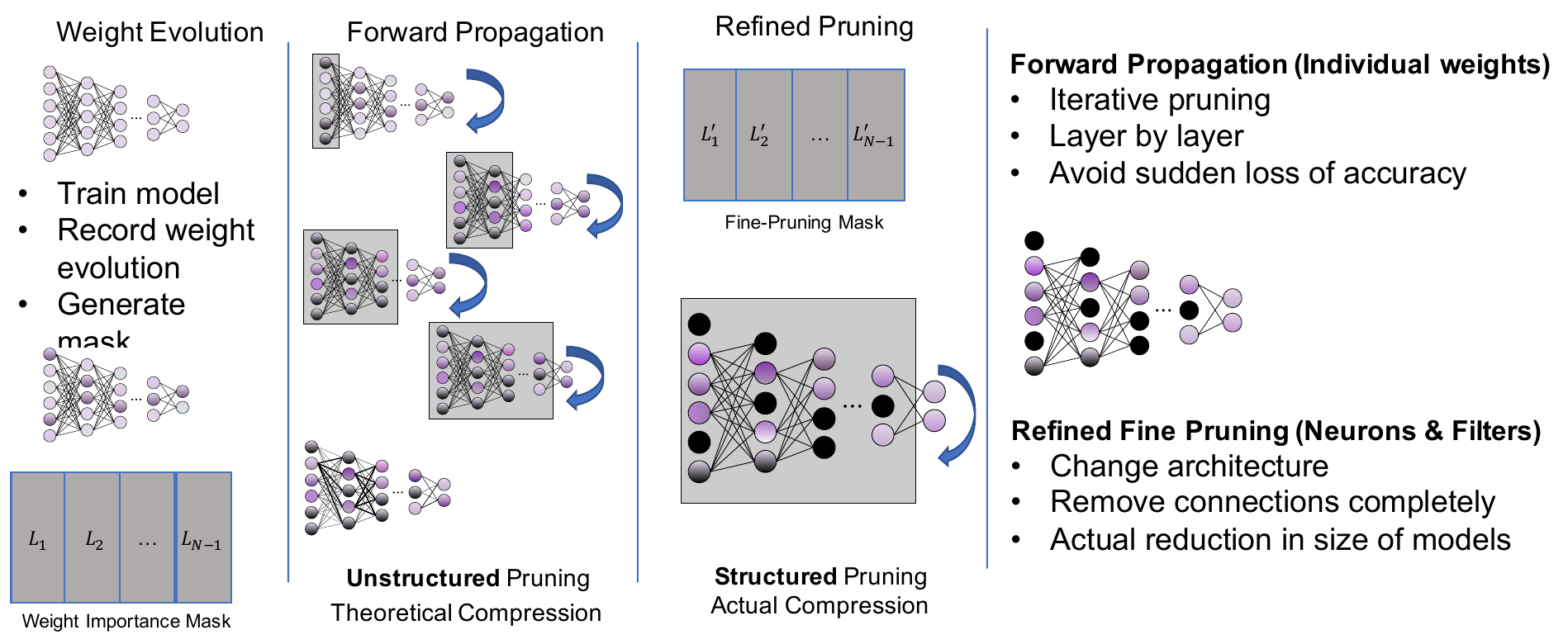}
\end{graphicalabstract}

\begin{highlights}
\item \textbf{Efficient Pruning for Large Language Models:} This research introduces a novel pruning method that evaluates the importance of individual weights throughout the training process, significantly optimizing model performance while reducing resource consumption.

\item \textbf{Impact of Compression on Model Performance:} Through comprehensive experiments, the study demonstrates that moderate pruning can enhance model efficiency, but excessive compression leads to substantial performance degradation in both language and multimodal models.

\item \textbf{Sustainable AI Development:} The findings emphasize the need for optimized AI models to reduce the environmental impact, addressing critical issues like carbon footprint, electricity, and water consumption associated with training and deploying large-scale AI systems.

\end{highlights}

\begin{keyword}
Large Language Models (LLMs) \sep Weight Evaluation \sep Model Pruning \sep Sustainable AI \sep Performance Optimization
\end{keyword}

\end{frontmatter}



\section{Introduction}

The rapid advancement of artificial intelligence, particularly in the development and deployment of large language models (LLMs) like ChatGPT, has brought about significant benefits across various domains. These models, with billions of parameters, have demonstrated unparalleled capabilities in tasks such as natural language understanding, translation, and text generation. However, the growing scale of these models comes with substantial challenges, particularly in terms of computational cost, environmental impact, and resource consumption, which underscore the need for optimization.

One of the most pressing concerns associated with training and deploying LLMs is their considerable carbon footprint. The computational resources required to train these models are immense, resulting in significant CO2 emissions. For instance, the training of BERT, a model with 110 million parameters \citep{Zhou}, is estimated to have a carbon footprint comparable to a round trip flight across the United States \citep{Strubell2020}. For larger models like ChatGPT, which contains 137 billion parameters \citep{Gooding2023}, the environmental impact is even more pronounced, equating to the annual CO2 emissions of 13,483 Americans \citep{Ludvigsen2022}. This alarming scale of emissions highlights the urgency of developing more sustainable approaches to AI. In addition to carbon emissions, the electricity consumption associated with LLMs is staggering. The daily global usage of ChatGPT alone consumes an estimated 12 million kilowatt-hours (KWh) of electricity, equivalent to the monthly electricity consumption of 90,000 Danish households \citep{towardsdatascienceChatGPTsElectricity}. This energy demand not only poses a significant operational cost but also contributes to the environmental burden, particularly in regions where electricity is predominantly generated from fossil fuels. Another critical aspect is the water footprint of AI models, a less commonly discussed but equally important factor. Training large models like ChatGPT requires vast amounts of water for cooling data centers, with estimates suggesting that training such a model can consume over 700,000 liters of freshwater, enough for producing 370 BMW cars or 320 Tesla electric vehicles \citep{Li2023}. Moreover, even a simple conversation with ChatGPT, involving 20-50 questions and answers, could necessitate the equivalent of a 500ml bottle of water \citep{Li2023}. This substantial water usage raises concerns, especially in areas facing water scarcity. Given the finite nature of natural resources and the increasing size of AI models, it is evident that the current trajectory is detrimental to the environmental stability. The growing computational power required to train and deploy these models exacerbates the strain on global resources, necessitating a reevaluation of how these models are developed and maintained. Optimization techniques, such as model pruning and efficient resource management, are crucial in addressing these challenges. By reducing the size of models without compromising their performance, we can mitigate the environmental impact, reduce operational costs, and make AI more sustainable in the long run.

\textbf{Contribution}

In this research, we introduce a novel approach to pruning large language models (LLMs) by systematically evaluating the performance of each individual weight throughout the training process. The remainder of the paper is structured as follows: Section \ref{sec:relWorks} reviews recent trends in LLM pruning, Section \ref{sec:propApp} details our proposed methodology, and Section \ref{sec:expRes} presents an overview of the performance of our approach. Finally, Section \ref{sec:limFut} concludes the paper by discussing the limitations of our work and potential directions for future research.

\section{Related Work}\label{sec:relWorks}
Magnitude pruning \citep{han2015learning} is a standard technique to induce sparsity in neural networks by removing individual weights based on their magnitudes, typically determined either locally within each layer or globally across the entire network. Despite its simplicity, it has been effective in finding extremely sparse networks \citep{Frankle2019} and is considered a strong baseline approach \citep{Blalock2020} for neural network sparsification. Dettmers et al. \citep{dettmers2022gpt3} observed emergent large magnitude features in Transformer-based large language models (LLMs), noting that when LLMs reach around 6B parameters, a small set of hidden state features emerges with significantly larger magnitudes than others, which are crucial for predictive performance. In the context of compressing recent LLMs, methods like LLM-Pruner \citep{ma2023llm} and FLAP \citep{an2024fluctuation} narrow network width by pruning coupled structures, while Sheared-LLaMA \citep{xia2023sheared} reduces both network width and depth by removing entire layers. Although pruning methods that incorporate both width and depth aspects exist \citep{xia2022structured, kurtic2024ziplm}, there remains a need for detailed analysis comparing these factors' impact on LLM inference efficiency. Traditional pruning in Deep Neural Networks (DNNs) faces unique challenges when applied to LLMs, which have a large number of parameters and require significant computational resources \citep{brown2020language}. Various pruning methods for LLMs fall into unstructured and structured categories. Unstructured pruning methods \citep{dong2017learning, chen2020tight, chen2021knowledge} set unimportant individual weights to zero, maintaining performance but resulting in sparse weight matrices that are less hardware-efficient. Methods like SparseGPT \citep{frantar2023sparsegpt} and Wanda \citep{sun2023simple} use sophisticated weight updates and pruning without retraining, respectively, while PST \citep{li2022parameter} combines unstructured pruning with efficient fine-tuning. Structured pruning methods \citep{chen2021only, chen2023otov2} remove entire groups of parameters, maintaining dense weight matrices and improving hardware efficiency. Techniques such as LLM-Pruner \citep{ma2023llm} and LoRAPrune \citep{zhang2023loraprune} focus on efficient deployment and inference acceleration, with Sheared-LLaMA \citep{xia2023sheared} aiming to prune models to a target architecture and train them dynamically. Furthermore, the compression of language models has garnered significant attention, leading to various methods like network pruning, knowledge distillation, and quantization \citep{bai2020binarybert, brown2020language, devlin2018bert}. Pruning, especially structural pruning, remains a crucial focus due to its hardware-friendly nature, with methods varying from l1-dependent pruning \citep{zafrir2021prune} to more advanced techniques like the optimal brain surgeon \citep{LeCun}. Efficient compression and low-resource strategies are increasingly essential, with advancements in layer-wise optimal brain surgeon and data-free pruning approaches aiming to optimize the balance between compression efficiency and training data availability \citep{kurtic2024ziplm, srinivas2015data}.

\section{Proposed Approach}\label{sec:propApp}
\begin{figure}[htbp]
    \centering
        \fbox{\includegraphics[scale=0.4]{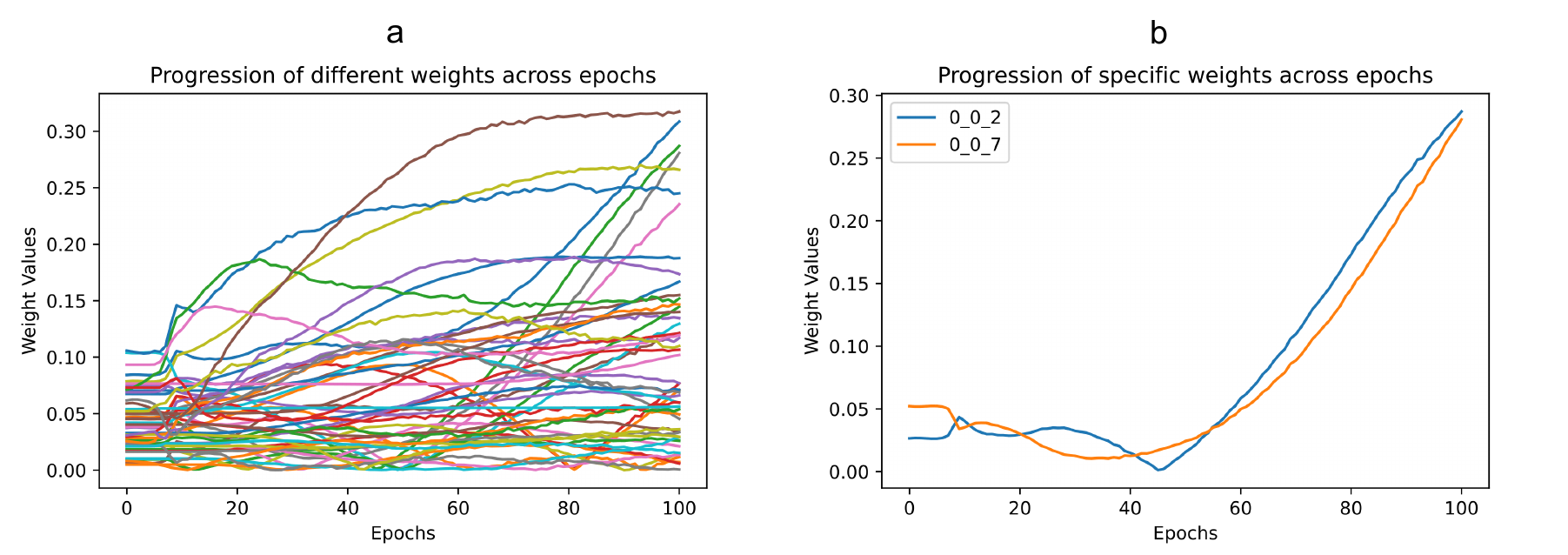}}
    \caption{(a) Depicts the evolution of randomly selected weights over 100 training epochs. \\(b) Illustrates the progression of specific weights across the same 100 training epochs.}
    \label{FIG:wt_progression_all_specific}
\end{figure}

\begin{figure}[htbp]
    \centering
    \includegraphics[width=0.9\textwidth]{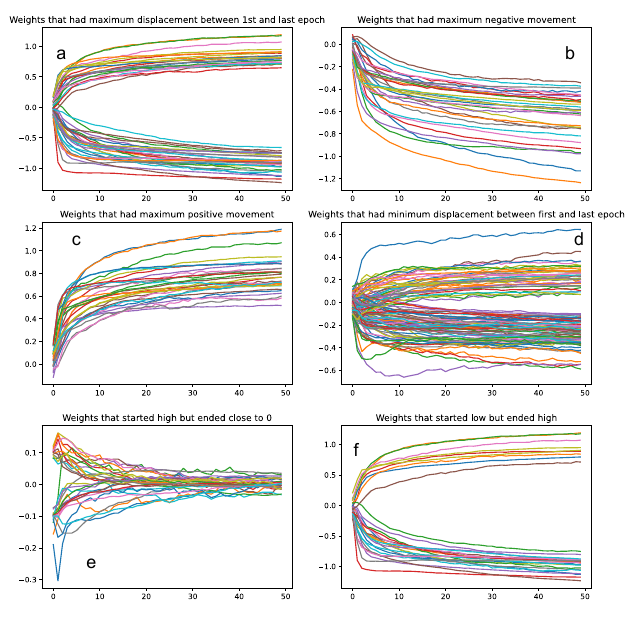}
    \caption{Evolution of weights over training epochs. Fig (a) and (b) show weights that increase dramatically, while (c) and (d) illustrate weights that fluctuate minimally. Fig (e) and (f) further demonstrate the divergence and stability of weights, emphasizing patterns critical for effective pruning.}
    \label{fig:EvolWeights}
\end{figure}

The central element of the pruning method proposed in this research lies in monitoring the evolution of parameters over time. This involves systematically observing how parameter values shift throughout the training process across multiple epochs. Figure \ref{FIG:wt_progression_all_specific} presents two visualizations of the network's weight dynamics: one tracks the changes in randomly selected weights over 100 epochs, while the other follows the trajectory of specific weights within the same training epochs. The initial weight values are chosen randomly and then adjusted during training as per the training algorithm. We note the overall shift of weights of the network, with some weights jumping high radically and some staying low over the training phase. This behavior is further illustrated in Figure \ref{fig:EvolWeights}, where the evolution of network weights across multiple epochs is displayed. Specifically, in subfigure \ref{fig:EvolWeights}a, we observe a subset of weights that increase dramatically early in the training process. In contrast, subfigure \ref{fig:EvolWeights}b highlights weights that remain relatively stable throughout training. Subfigure \ref{fig:EvolWeights}c shows a mix of these behaviors, with some weights fluctuating while others stabilize. 

Subfigure \ref{fig:EvolWeights}d demonstrates the long-term evolution, emphasizing the divergence between weights that become highly influential and those that remain low, indicating their potential for pruning. Furthermore, subfigures \ref{fig:EvolWeights}e and \ref{fig:EvolWeights}f offer additional insights into the final distribution of weights, illustrating both the stability of low-magnitude weights and the radical shifts in others, further reinforcing the potential for targeted pruning to optimize the model.

Our approach incorporates a system that assigns weights to the parameter magnitudes, emphasizing those values closer to the end of the training epochs while still considering earlier data. The importance of each parameter is calculated by multiplying its magnitude by a corresponding weight and then averaging these values, which helps in constructing an importance vector that maps the evolution of parameters throughout the training epochs.

\begin{table}[h!]
\caption{To assess the significance of parameters over training epochs, we log the value of each parameter at the end of each epoch, organizing these values into columns. The aggregated significance is obtained by performing a weighted sum of each weight’s magnitude. The multipliers in the bottom row show the extent to which each magnitude is considered in the calculation.}
\begin{tabular}{|p{0.16\linewidth} | p{0.08\linewidth} | p{0.08\linewidth} | p{0.08\linewidth} | p{0.08\linewidth} | p{0.18\linewidth}|}

\toprule
Weight \# & Epoch 1 &Epoch 2 & Epoch ...& Epoch k& Aggregated Importance\\
\midrule
1& 4& 6& ...&3& 14\\
2& 8& 9& ...&5& 13\\
3& 6& 8& ...&8& 6.66\\
4& 2& 5& ...&9& 4.66\\
\textbf{Multiplier}& $*$1 & $*$2 &...& $*$k & \\

\bottomrule
\end{tabular}
\label{TBL:Aggregate}
\end{table}

For example, to calculate the weighted importance of a particular weight or filter, we compile a log of its magnitude recorded at each epoch during training. This log allows us to assess the weighted significance according to the provided equation. The resulting score represents the importance of a parameter in terms of its magnitude and how it has changed throughout the training process. Table \ref{TBL:Aggregate} displays the magnitude logs for various weights across epochs. Using our method, the most significant weights were found to be Weight 1 (with a score of 14), Weight 2 (with a score of 13), and Weight 3 (with a score of 6.66).

To generalize this calculation, we define a vector for each weight or filter (\(val_{i} = [val_{i1}, val_{i2}, val_{i3},... val_{in}]\)), where each entry corresponds to the weight’s magnitude at a specific epoch across the \(n\) training epochs. This vector forms the basis for calculating the weighted significance using the following equation, which prioritizes the most recent \(k\) epochs:

\begin{equation} Imp_{i}=\frac{\sum_{L=0}^{k} val_{i(n-L)}*(n-L)}{\sum_{L=0}^{k} (n-L)} \end{equation}

Here, \(L\) varies from 0 to \(k\), with 0 representing the most recent epoch and \(k\) counting back from the final epoch. The resulting importance matrix becomes a crucial tool for assessing weight significance and guiding the network pruning strategy.

\section{Experiment And Results}\label{sec:expRes}

To check the consistency of our methods, two key experiments were conducted. These experiments focused on evaluating the effects of pruning, a process that reduces the number of parameters in a model, on model performance. The first experiment tested a scaled-down LLM trained from scratch, while the second involved a large pre-trained multimodal model. Both experiments aimed to determine how much compression could be applied to these models before significant performance degradation occurred. Before looking at the individual experiments, we take a look at the general procedure.

\subsection{Record Weighted Average}
In addition to directly training the model, a cloned version is maintained alongside it. The parameters of this clone are updated through a weighted average method that integrates historical parameter values across the training epochs. Initially, the cloned model's parameters are set to zero before the training starts. After each training step, both the original model's parameters and the corresponding parameters in the clone are updated. The updated values in the clone are computed as a weighted average, combining the existing parameters with the new ones from the original model, based on the current epoch. This approach ensures that recent updates are given more significance in the clone. The weighted average process, which gradually incorporates the model's parameter values over the epochs, is expressed as:

\begin{equation}
q_{\text{new}} = \frac{q_{\text{old}} \times S_{\text{prev}} + p \times (n + 1)}{S}
\end{equation}

Where:
\begin{itemize}
    \item \( q_{\text{new}} \) are the updated parameters in the cloned model.
    \item \( q_{\text{old}} \) are the previous parameters in the cloned model.
    \item \( p \) are the current parameters in the original model.
    \item \( n \) is the current epoch number.
    \item \( S_{\text{prev}} \) is the sum of weights from epoch 1 to \( n \) (inclusive).
    \item \( S \) is the sum of weights from epoch 1 to \( n+1 \) (inclusive).
\end{itemize}

\subsection{Model Training and Pruning}
\begin{itemize}
    \item Step I: The Transformer model is trained over 5000 epochs, with weight changes recorded throughout the process.
    \item Step II: After training, the model undergoes pruning based on the weighted parameters, followed by an additional 50 epochs of fine-tuning to maintain effective compression:
    \begin{itemize}
        \item The importance of each parameter is assessed by evaluating its weighted absolute value:
        \begin{equation}
      W_{\text{abs}} = |W|
      \end{equation}
      \item A pruning threshold is determined by scaling the standard deviation of these absolute values with a specific rate:
      \begin{equation}
   \text{Threshold} = \sigma(W_{\text{abs}}) \times \text{Prune Rate}
   \end{equation}
   Here, \( \sigma(W_{\text{abs}}) \) represents the standard deviation of \( W_{\text{abs}} \), and "Prune Rate" is a constant that dictates the extent of pruning.
   \item Parameters falling below the pruning threshold are set to zero:
   \begin{equation}
   P = 
   \begin{cases} 
   0 & \text{if } W_{\text{abs}} < \text{Threshold} \\
   P & \text{otherwise}
   \end{cases}
   \end{equation}
    \end{itemize}
\end{itemize}
The effectiveness of Step II is evaluated by varying the pruning rates and observing the corresponding loss values.

\subsection{Experiment 1: Scaled-Down LLM}

\subsubsection{Model and Dataset:}
The first experiment utilized a scaled-down version of a ChatGPT-like Large Language Model (LLM), consisting of 10,788,929 (10.7 million) adjustable parameters. This substantial language model is founded on the Transformer architecture, which is specialized in generating text for natural language processing tasks \citep{vaswani2017attention}. The model comprises several Transformer blocks, each containing the following components:

\begin{itemize}
    \item \textbf{Embedding Layer}: Input tokens are transformed into 384-sized vectors, forming the initial layer input.
    \item \textbf{Multi-Head Self-Attention Mechanism}: This mechanism, with 6 heads, directs focus across different segments of the input sequence using linear transformations without bias, scaling the attention scores by the square root of the embedding size (384).
    \item \textbf{Positional Embeddings}: Positional embeddings of the same size (384) are added to the input tokens to signify sequence positions, enabling the model to recognize word order.
    \item \textbf{Feed-Forward Network}: Following the attention mechanism, each Transformer block includes a feed-forward network with two linear layers and a hidden layer size of 1536, activated by ReLU. Dropout is applied with a rate of 0.0 to reduce overfitting.
    \item \textbf{Layer Normalization}: This process standardizes activations across features and enhances learning, applied after both the self-attention and feed-forward network layers.
    \item \textbf{Final Layer}: The final layer performs a linear transformation, correlating output embeddings with the vocabulary size based on the unique input text characters.
\end{itemize}

The complete model stacks 6 of these Transformer blocks, processing the input independently to grasp varying semantic layers in the text. The model was trained from scratch using the complete works of Shakespeare, which served as the training dataset. The primary task for this model was next-token prediction, a common benchmark task in language modeling. All 10,788,929 trainable parameters were tracked throughout the training epochs.
\subsubsection{Training Procedure:}
The model was subjected to a series of pruning tests, where the parameter count was reduced incrementally. The compression levels ranged from 0\% (no pruning) to 94\% (high pruning). The effect of each level of compression on model performance was monitored by tracking the loss associated with the next-token prediction task.

\subsection{Experiment 2: Multimodal Model}
\subsubsection{Model and Dataset:}
The second experiment centered around a pre-trained Phi-3-vision model, a large multimodal model with 4.2 billion parameters, designed for both language and vision tasks. The dataset for this experiment comprised a variety of Burberry products, categorized under different product types such as hats, gloves, and sunglasses. The dataset included both textual descriptions and corresponding images of the products.

\subsubsection{Training Procedure:}
In this experiment, the Phi-3-vision model \citep{abdin2024phi} was fine-tuned for 10 epochs using the Burberry dataset \citep{huggingfaceDBQBurberryProductpricesUnitedStatesDatasets}. After fine-tuning, the model underwent pruning at various compression levels, similar to the first experiment. The performance of the model was evaluated by measuring the Mean Absolute Error (MAE) at each level of compression to understand the impact of pruning on its ability to process and classify multimodal inputs.

\subsection{Results}
\subsubsection{Results of Experiment 1}
The results from the first experiment, as shown in Table \ref{tab:compression_loss}, indicate that the model could tolerate compression up to 60\% without a significant increase in loss. At this level, the compression loss was recorded at 1.656, which was lower than the initial loss of 1.9. However, further compression led to an increase in loss, with a sharp rise observed at 70\% compression and above, where the loss eventually escalated to 3.098 at 94\% compression. This demonstrates that moderate pruning can lead to a reduction in loss, but excessive pruning severely hampers model performance.

\begin{table}[h!]
\centering
\begin{tabular}{|c|c|}
\hline
\textbf{Compression (\%)} & \textbf{Loss} \\
\hline
0 & 1.900 \\
0.1 & 1.977 \\
0.2 & 1.932 \\
0.3 & 1.864 \\
0.4 & 1.782 \\
0.5 & 1.693 \\
0.6 & 1.656 \\
0.7 & 1.747 \\
0.8 & 2.133 \\
0.9 & 2.703 \\
0.94 & 3.098 \\
\hline
\end{tabular}
\caption{Compression Loss in Experiment 1}
\label{tab:compression_loss}
\end{table}

\begin{figure}[htbp]
    \centering
    \includegraphics[width=\textwidth]{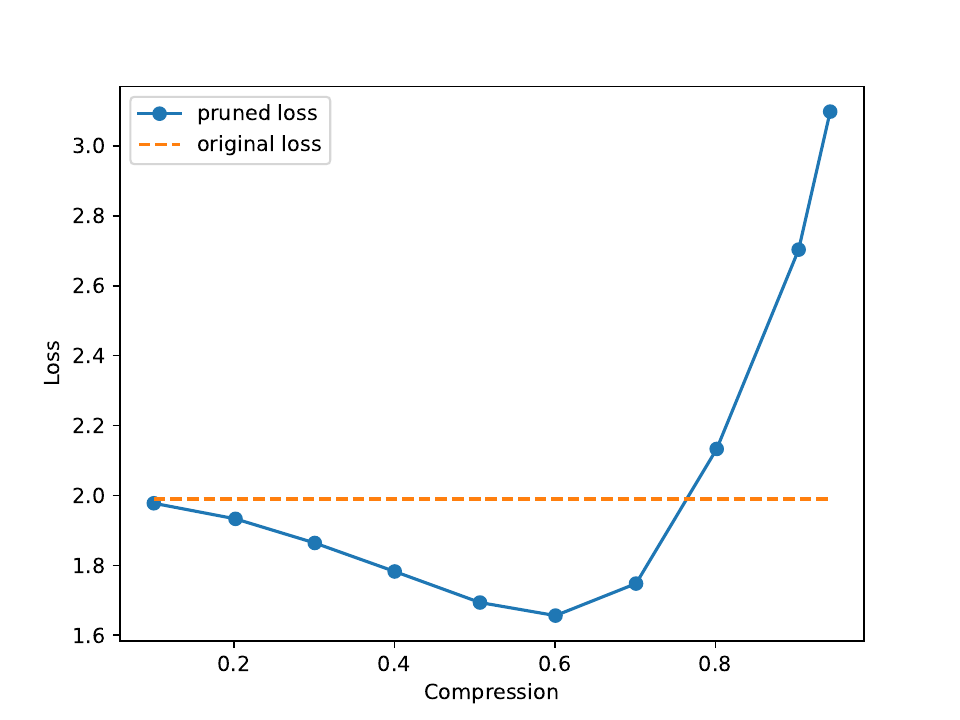}
    \caption{Loss as a function of compression levels. The figure shows a decrease in loss up to 60\% compression, after which a sharp increase is observed, particularly beyond 70\% compression, consistent with the trends detailed in Table \ref{tab:compression_loss}.}
    \label{fig:loss_vs_prune}
\end{figure}

The impact of compression on model loss is further visualized in Figure \ref{fig:loss_vs_prune}. The figure illustrates how the loss changes as the model undergoes increasing levels of compression. As shown, the loss decreases slightly up to a compression level of around 60\%, corroborating the trend observed in Table \ref{tab:compression_loss}, where a reduction in loss was recorded at moderate compression levels. However, beyond this point, the figure reveals a steep increase in loss, especially as compression exceeds 70\%, culminating in a significant rise at 94\% compression. This visual representation reinforces the findings from the table, indicating that while some pruning can reduce loss, excessive pruning significantly impairs the model's ability to perform, leading to a sharp increase in loss.

\subsubsection{Results of Experiment 2}
In the second experiment, the Phi-3-vision model exhibited stable performance at lower compression levels, as detailed in Table \ref{tab:mae}. The MAE decreased from 439 at 0\% compression to 374 at 10\% compression, suggesting that some level of pruning can improve the model’s performance, possibly by eliminating redundant parameters. However, at compression levels beyond 25\%, the MAE began to rise again, and a drastic increase was observed at 48\% compression, where the MAE surged to 11,041. This indicates that while the model can handle moderate pruning, aggressive pruning beyond 30\% significantly deteriorates its performance.

\begin{table}[h!]
\centering
\begin{tabular}{|c|c|}
\hline
\textbf{Compression (\%)} & \textbf{MAE} \\
\hline
0 & 439 \\
5 & 380 \\
10 & 374 \\
12 & 378 \\
15 & 384 \\
20 & 397 \\
25 & 457 \\
30 & 401 \\
37 & 474 \\
48 & 11041 \\
59 & 950 \\
67 & 961 \\
76 & 961 \\
\hline
\end{tabular}
\caption{MAE in Experiment 2}
\label{tab:mae}
\end{table}

\begin{figure}[htbp]
    \centering
    \includegraphics[width=\textwidth]{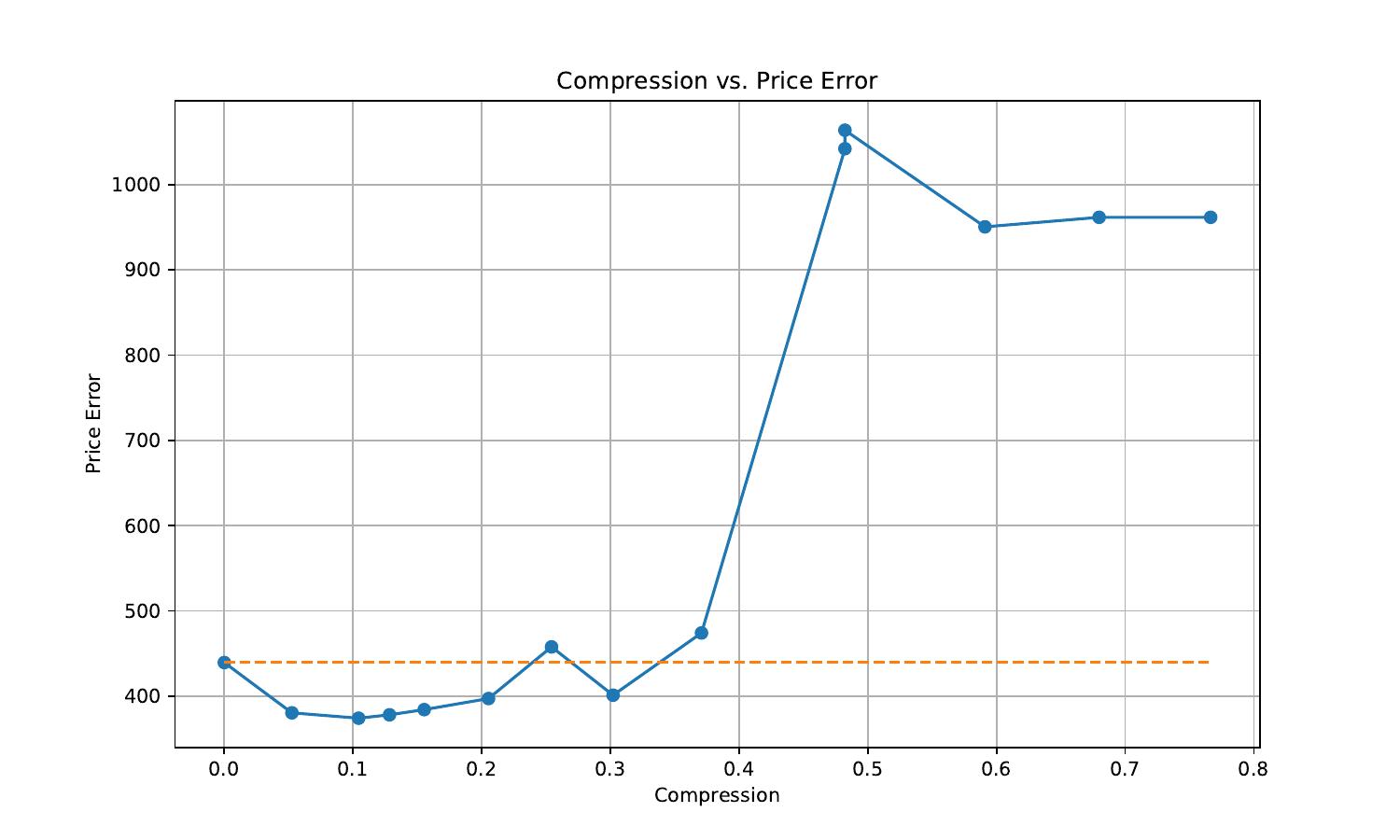}
    \caption{Price error as a function of compression levels. The figure demonstrates that while the model maintains a relatively low error up to moderate compression levels, the error escalates sharply beyond 30\% compression, consistent with the MAE trends observed in Table \ref{tab:mae}.}
    \label{fig:comp_vs_error}
\end{figure}

The relationship between compression levels and model error is further illustrated in Figure \ref{fig:comp_vs_error}. The figure visualizes how the model's price prediction error varies with increasing compression levels. As shown, the model maintains a relatively low error up to moderate compression levels, reinforcing the findings in Table \ref{tab:mae} where the MAE decreases slightly with initial pruning. However, as the compression surpasses 30\%, the error begins to escalate, culminating in a significant spike at higher compression levels. This sharp increase at 48\% compression aligns with the drastic rise in MAE observed in the table, indicating that the model's capacity to handle pruning is limited and aggressive compression can severely impair its predictive performance.

\section{Limitations And Future Work}\label{sec:limFut}

While the research on pruning and optimizing Large Language Models (LLMs) shows promising results, it is essential to recognize certain limitations that present challenges for future development.

\begin{itemize}
    \item \textbf{Specialized Use Cases:} While fine-tuning LLMs for specific purposes can yield high performance, it may limit the model's applicability across a broader range of tasks, requiring more adaptable solutions.
    \item \textbf{Adaptation to Model Size:} As LLMs grow larger, the proportion of parameters that can be effectively pruned decreases unless more advanced techniques are applied, highlighting the need for methods that can handle large-scale models efficiently.
    \item \textbf{Memory Considerations:} Managing the memory requirements for storing a weighted average of parameters in models with millions or billions of parameters presents a significant challenge, necessitating memory-efficient strategies.
\end{itemize}

\subsection{Future Work}

Looking forward, the future work in this area is both hopeful and exciting. There is a strong potential for tangible energy savings by optimizing LLMs more efficiently, making AI systems not only faster but also more sustainable. The research community is also poised to explore smarter pruning methods that could overcome current limitations, enabling deeper compression without sacrificing model accuracy or generalization capabilities. As we continue to push the boundaries of LLMs, the focus will be on balancing innovation with sustainability, ensuring that the advancements in AI contribute positively to both technological progress and environmental responsibility.

\bibliographystyle{elsarticle-harv} 






\end{document}